\documentclass[conference,letterpaper]{IEEEtran}

\IEEEoverridecommandlockouts

\usepackage{cite}
\usepackage{amsmath,amssymb,amsfonts}
\usepackage{algorithmic}
\usepackage{graphicx}
\usepackage{textcomp}
\usepackage{xcolor}
\usepackage{booktabs}
\usepackage{pgfplots}
\pgfplotsset{compat=1.18}

\def\BibTeX{{\rm B\kern-.05em{\sc i\kern-.025em b}\kern-.08em
T\kern-.1667em\lower.7ex\hbox{E}\kern-.125emX}}
\begin{document}

\title{Multilingual Financial Fraud Detection Using Machine Learning and Transformer Models: A Bangla–English Study}

\author{
\IEEEauthorblockN{
Mohammad Shihab Uddin\IEEEauthorrefmark{1},
Md Hasibul Amin\IEEEauthorrefmark{2}\thanks{*Corresponding author: hasibulamin.cse@gmail.com},
Nusrat Jahan Ema\IEEEauthorrefmark{1},
Bushra Uddin\IEEEauthorrefmark{1}, \\
Tanvir Ahmed\IEEEauthorrefmark{1}, and  
Arif Hassan Zidan\IEEEauthorrefmark{1}
}

\IEEEauthorblockA{
\IEEEauthorrefmark{1}Augusta University, Augusta, GA 30912, USA
}

\IEEEauthorblockA{
\IEEEauthorrefmark{2}University of Houston, Houston, TX 77004, USA
}
}

\maketitle

\begin{abstract}
Financial fraud detection has emerged as a critical research challenge amid the rapid expansion of digital financial platforms. Although machine learning approaches have demonstrated strong performance in identifying fraudulent activities, most existing research focuses exclusively on English-language data, limiting applicability to multilingual contexts. Bangla (Bengali), despite being spoken by over 250 million people, remains largely unexplored in this domain. In this work, we investigate financial fraud detection in a multilingual Bangla–English setting using a dataset comprising legitimate and fraudulent financial messages. We evaluate classical machine learning models (Logistic Regression, Linear SVM, and Ensemble classifiers) using TF-IDF features alongside transformer-based architectures. Experimental results using 5-fold stratified cross-validation demonstrate that Linear SVM achieves the best performance with 91.59\% accuracy and 91.30\% F1 score, outperforming the transformer model (89.49\% accuracy, 88.88\% F1) by approximately 2 percentage points. The transformer exhibits higher fraud recall (94.19\%) but suffers from elevated false positive rates. Exploratory analysis reveals distinctive patterns: scam messages are longer, contain urgency-inducing terms, and frequently include URLs (32\%) and phone numbers (97\%), while legitimate messages feature transactional confirmations and specific currency references. Our findings highlight that classical machine learning with well-crafted features remains competitive for multilingual fraud detection, while also underscoring the challenges posed by linguistic diversity, code-mixing, and low-resource language constraints.
\end{abstract}


\section{Introduction}

The rapid growth of digital financial services, including online banking, mobile payment platforms, and electronic commerce, has significantly increased the scale and complexity of financial transactions. While these technologies improve efficiency and accessibility, they also introduce new opportunities for fraudulent activities such as phishing scams, deceptive financial advertisements, and fraudulent transaction descriptions. Financial fraud continues to cause substantial economic losses worldwide, making automated fraud detection an essential research problem for modern financial systems.

Machine learning and natural language processing (NLP) techniques have been widely adopted to detect fraudulent behavior from transactional records and text-based financial data. 
 Prior studies have demonstrated that textual features extracted from transaction descriptions, customer messages, and online financial communications can effectively capture fraud-related patterns. However, most existing financial fraud detection systems are developed and evaluated primarily on English-language datasets, which limits their applicability in multilingual environments.

In many real-world settings, particularly in South Asia, financial communication frequently involves multiple languages or mixed-language usage. Bangla (Bengali), spoken by over 250 million people worldwide, is commonly used alongside English in digital financial platforms, informal transaction messages, and online advertisements. Despite its widespread usage, Bangla remains a low-resource language in NLP, especially for domain-specific tasks such as financial fraud detection and prior research shows that model performance often correlates strongly with available training resources in low-resource languages \cite{joshi2020state, artetxe2020cross, alam2021review}. The scarcity of labeled datasets, complex morphology, and frequent code-mixing with English pose additional challenges that are not addressed by English-centric models \cite{alam2021review, joshi2020state}.

Recent advances in deep learning, particularly transformer-based architectures, have significantly improved performance across a wide range of NLP tasks by enabling deep contextual representation learning through self-attention mechanisms \cite{vaswani2017attention,devlin2019bert,electronics14030554}. Multilingual transformer models further allow shared representations across languages and have demonstrated strong cross-lingual transfer capabilities across diverse linguistic settings \cite{conneau2020unsupervised,artetxe2019massively,pires2019multilingual}. Nevertheless, the effectiveness of these models for financial fraud detection in Bangla--English multilingual data remains largely unexplored.

This paper aims to address this gap by investigating financial fraud detection in a multilingual Bangla--English setting. We evaluate classical machine learning approaches using TF-IDF features as well as transformer-based architectures to analyze their performance across languages. Our study provides empirical insights into the challenges posed by linguistic diversity, data sparsity, and code-mixed text, and highlights the importance of multilingual modeling for robust financial fraud detection in low-resource environments.

The main contributions of this work are summarized as follows:
\begin{itemize}
    \item We study financial fraud detection in a multilingual Bangla--English context, an under-explored research area.
    \item We compare classical machine learning models with transformer-based architectures for multilingual fraud detection.
    \item We analyze the impact of linguistic characteristics and low-resource constraints on model performance.
\end{itemize}

\section{Related Work}

This section reviews prior research to position the present work and identify gaps motivating multilingual financial fraud detection.

\subsection{Financial Fraud Detection}

Financial fraud detection has long been important due to substantial losses across banking and digital payments. Early rule-based and statistical methods \cite{phua2010comprehensive,ferdus2025stock,ferdus2025does} often failed to generalize to evolving fraud patterns. Supervised learning algorithms have since demonstrated strong performance across diverse domains \cite{raju2024ontological,uddin2024epistemological,sabuj2024recondite}, with classical methods showing strong predictive performance for fraud detection \cite{ngai2011application,bahnsen2015cost,carcillo2018scarff,olowookere2020framework,sultana2024analyzing}. Studies have also addressed class imbalance \cite{dal2015adversarial,wang2025explainability} and evaluation bias \cite{dal2017credit}. However, most research focuses on structured numerical records and assumes monolingual English data.

\subsection{Text-Based Fraud Detection}

NLP applications for fraud detection use bag-of-words, n-grams, and TF-IDF representations \cite{salton1988term,rodriguez2022natural,boulieris2024fraud,craja2020deep}, with classical classifiers remaining widely used \cite{joachims1998text,fan2008liblinear,paper1}. Deep learning and transformer-based models \cite{11103694,vaswani2017attention,devlin2019bert} demonstrate state-of-the-art performance in detecting subtle linguistic patterns \cite{amin2022detecting,ZIDAN20251128}, though most studies focus exclusively on English datasets.

\subsection{Multilingual and Cross-Lingual NLP}

Multilingual NLP enables cross-lingual transfer \cite{joshi2020state,hossain2024advancements,jain2024knowledge}, with specialized Indic language models improving South Asian performance \cite{kakwani2020indicnlpsuite,khanuja2021muril}. Multilingual transformers learn language-agnostic representations \cite{conneau2020unsupervised,10459734}, encoding both universal and language-dependent features \cite{pires2019multilingual} and supporting zero-shot transfer \cite{artetxe2019massively}. However, performance degrades for low-resource languages \cite{joshi2020state,parvez2025can}.

\subsection{Code-Mixed and Low-Resource Language Challenges}

Real-world multilingual communication involves code-mixing and transliteration, with benchmarks highlighting modeling difficulties \cite{aguilar2020lince}. Low-resource languages pose challenges due to limited annotated datasets \cite{joshi2020state}.

\subsection{Bangla Language Processing}

Bangla remains underrepresented despite being widely spoken. Studies focus on sentiment analysis and text classification \cite{alam2021review,rahman2019benchmark,ahmed2022sentiment}, while domain-specific applications like financial fraud detection remain unexplored.

\subsection{Research Gap}

Key limitations include: (1) fraud detection focusing on structured data; (2) text-based studies using only English; (3) multilingual NLP rarely addressing domain-specific applications; (4) Bangla lacking domain-specific benchmarks. The success of ML-driven systems \cite{sultana2024deciphering,imam2024ml,islam2024hierarchical,arefin2024retail} motivates this study evaluating classical ML and transformer models for multilingual Bangla-English fraud detection.

\section{Methodology}

This section describes the feature extraction and classification models for financial fraud detection in a multilingual Bangla--English setting.

\subsection{Term Frequency--Inverse Document Frequency (TF-IDF)}

TF-IDF is a widely used feature representation technique that remains a strong baseline in text classification \cite{salton1988term,joachims1998text,arefin2024enhancing}. It transforms textual data into numerical vectors by weighting terms based on document frequency and corpus distribution \cite{salton1988term}. The TF-IDF weight for term $t$ in document $d$ is:
\begin{equation}
TF\text{-}IDF(t, d) = TF(t, d) \times IDF(t)
\end{equation}

We apply TF-IDF to both Bangla and English text using unigram and bigram representations to capture local contextual patterns indicative of fraud.

\subsection{Classical Machine Learning Algorithms}

We employ classical supervised algorithms suited for sparse TF-IDF features: Logistic Regression and Support Vector Machines, selected for their efficiency and strong generalization in text classification \cite{fan2008liblinear,joachims1998text}.

\subsubsection{Logistic Regression}
Logistic Regression estimates class probability given a feature vector $\mathbf{x} \in \mathbb{R}^n$:
\begin{equation}
P(y=1|\mathbf{x}) = \sigma(\mathbf{w}^\top \mathbf{x} + b)
\end{equation}
where $\mathbf{w}$ is the weight vector, $b$ is the bias, and $\sigma(\cdot)$ is the sigmoid function. Parameters are optimized via binary cross-entropy minimization.

\subsubsection{Support Vector Machine}
Linear SVMs find an optimal hyperplane maximizing the margin between classes using hinge loss, offering robustness to noise in sparse, high-dimensional feature spaces.

\subsection{Transformer-Based Architecture}

Transformers form the foundation of state-of-the-art NLP systems \cite{vaswani2017attention,devlin2019bert,conneau2020unsupervised}, processing sequences in parallel to capture long-range dependencies efficiently.

Input text is tokenized using subword tokenization, with embeddings combined with positional encodings. Each encoder layer applies multi-head self-attention followed by position-wise feed-forward networks, with residual connections and layer normalization for training stability.

In multilingual settings, shared subword vocabularies enable cross-language knowledge transfer---particularly beneficial for Bangla--English text with code-mixing and transliteration.

\section{Results \& Discussion}
\subsection{Exploratory Data Analysis}
To understand the characteristics of the multilingual financial fraud dataset~\cite{ahmed2025financial}, we analyze message length, language composition, discriminative features, and structural indicators across legitimate (ham) and fraudulent (scam) classes.

\subsubsection{Message Length Distribution}
\begin{figure}[htbp]
\centerline{\includegraphics[width=\columnwidth]{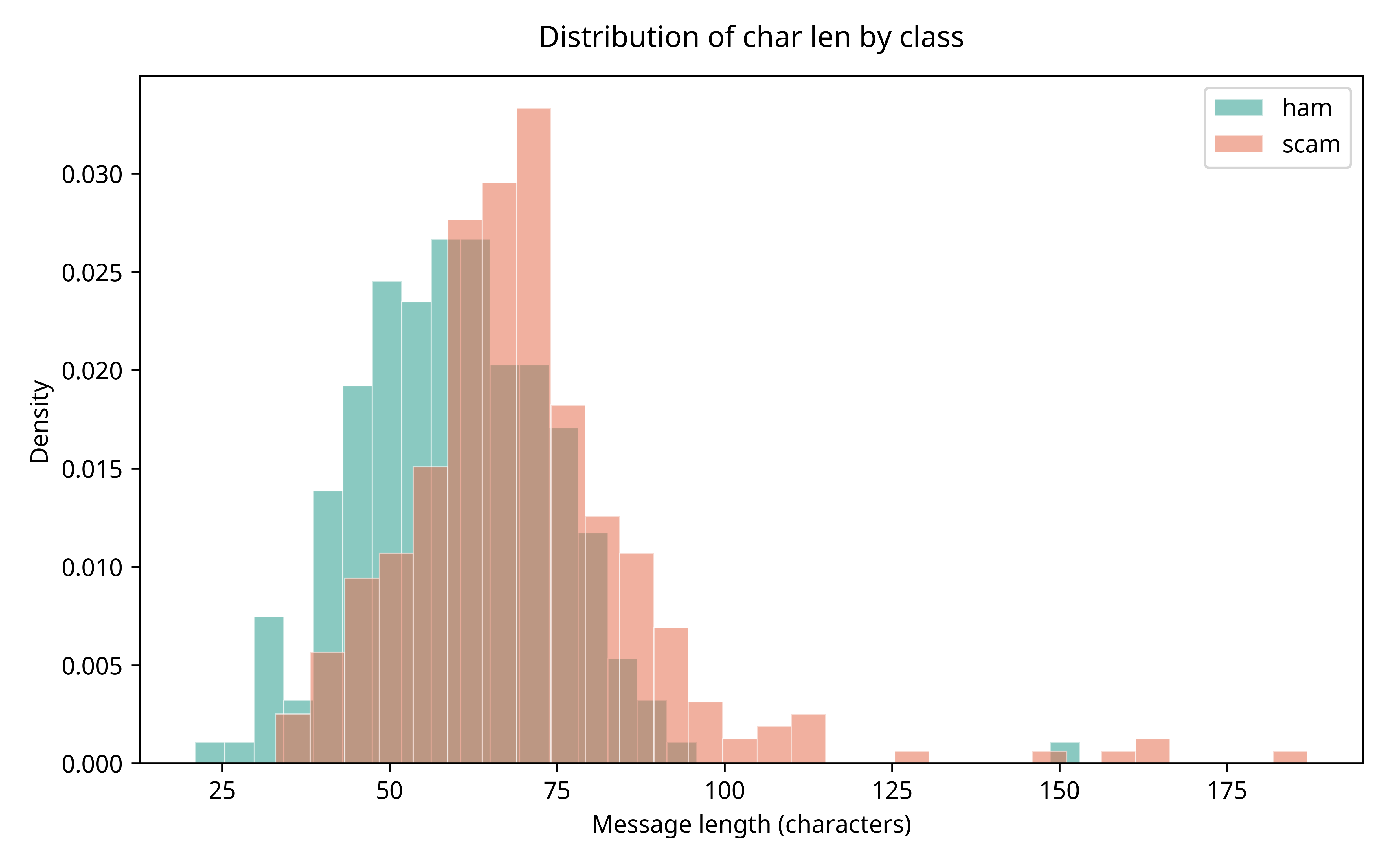}}
\caption{Distribution of message length (characters) by class.}
\label{fig:char_len}
\end{figure}

Fig.~\ref{fig:char_len} shows that scam messages tend to be longer than legitimate messages, peaking around 70--75 characters compared to 50--60 for ham, with some exceeding 150 characters. This suggests fraudulent messages contain more elaborate persuasive content.

\subsubsection{Language Composition}
\begin{figure}[htbp]
\centerline{\includegraphics[width=\columnwidth]{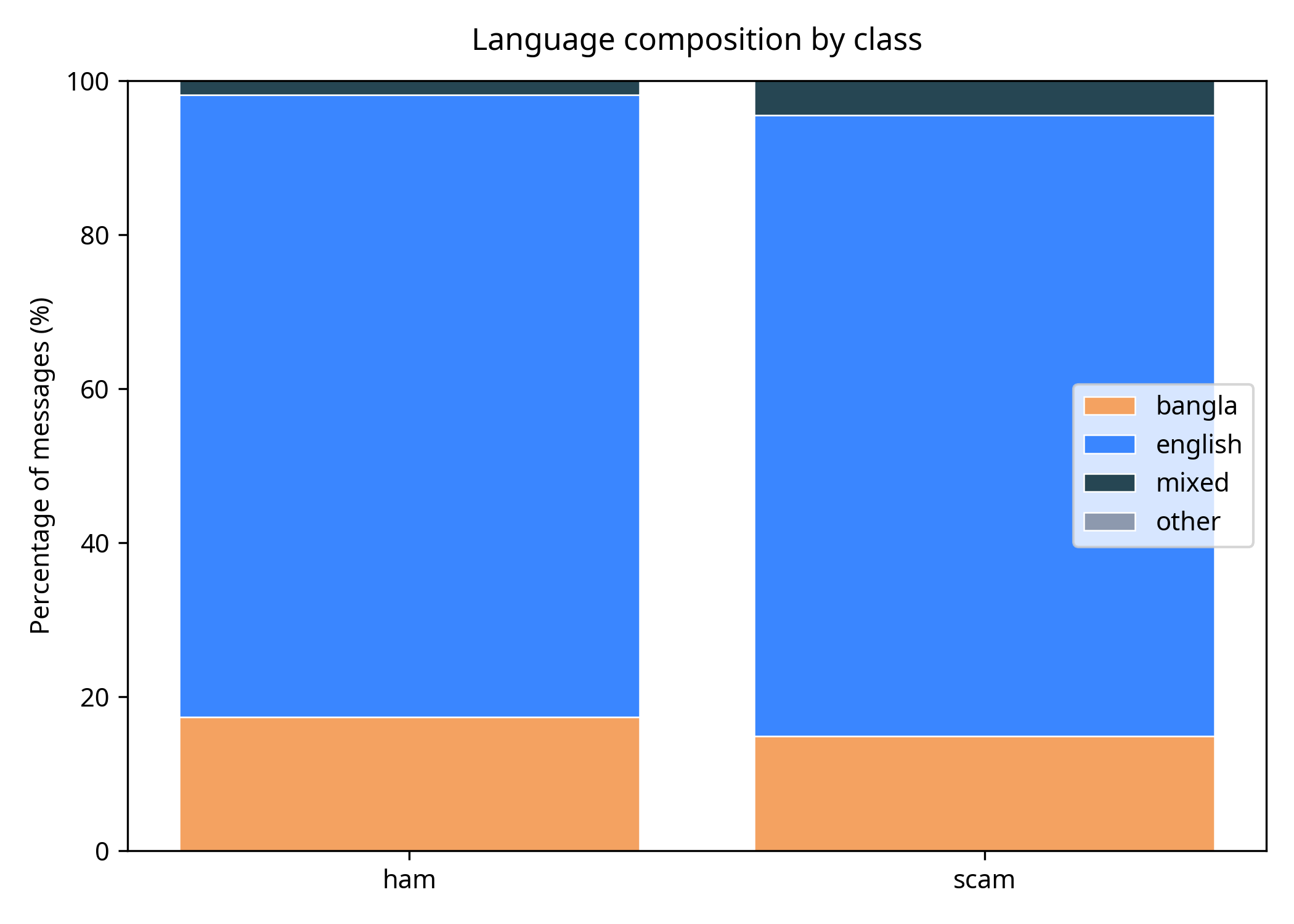}}
\caption{Language composition by class.}
\label{fig:lang_comp}
\end{figure}

Fig.~\ref{fig:lang_comp} shows English dominates both classes (around 80\%), with Bangla comprising 15--17\% and code-mixed messages forming a small but notable portion. This multilingual content underscores the need for models capable of handling linguistic diversity.

\subsubsection{Discriminative Lexical Features}
\begin{figure}[htbp]
\centerline{\includegraphics[width=\columnwidth]{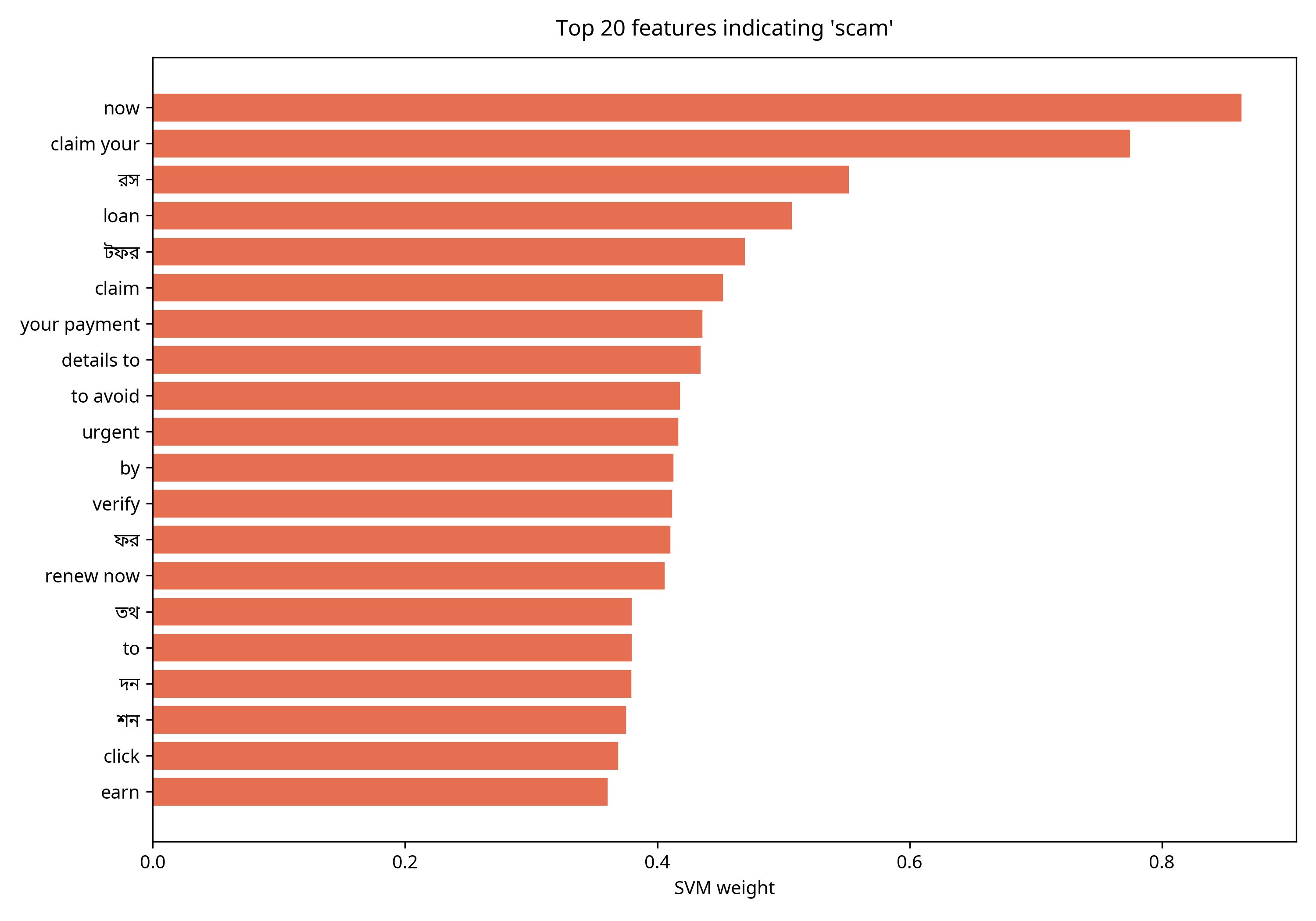}}
\caption{Top 20 TF-IDF features with highest SVM weights indicating the scam class.}
\label{fig:top_scam}
\end{figure}

\begin{figure}[htbp]
\centerline{\includegraphics[width=\columnwidth]{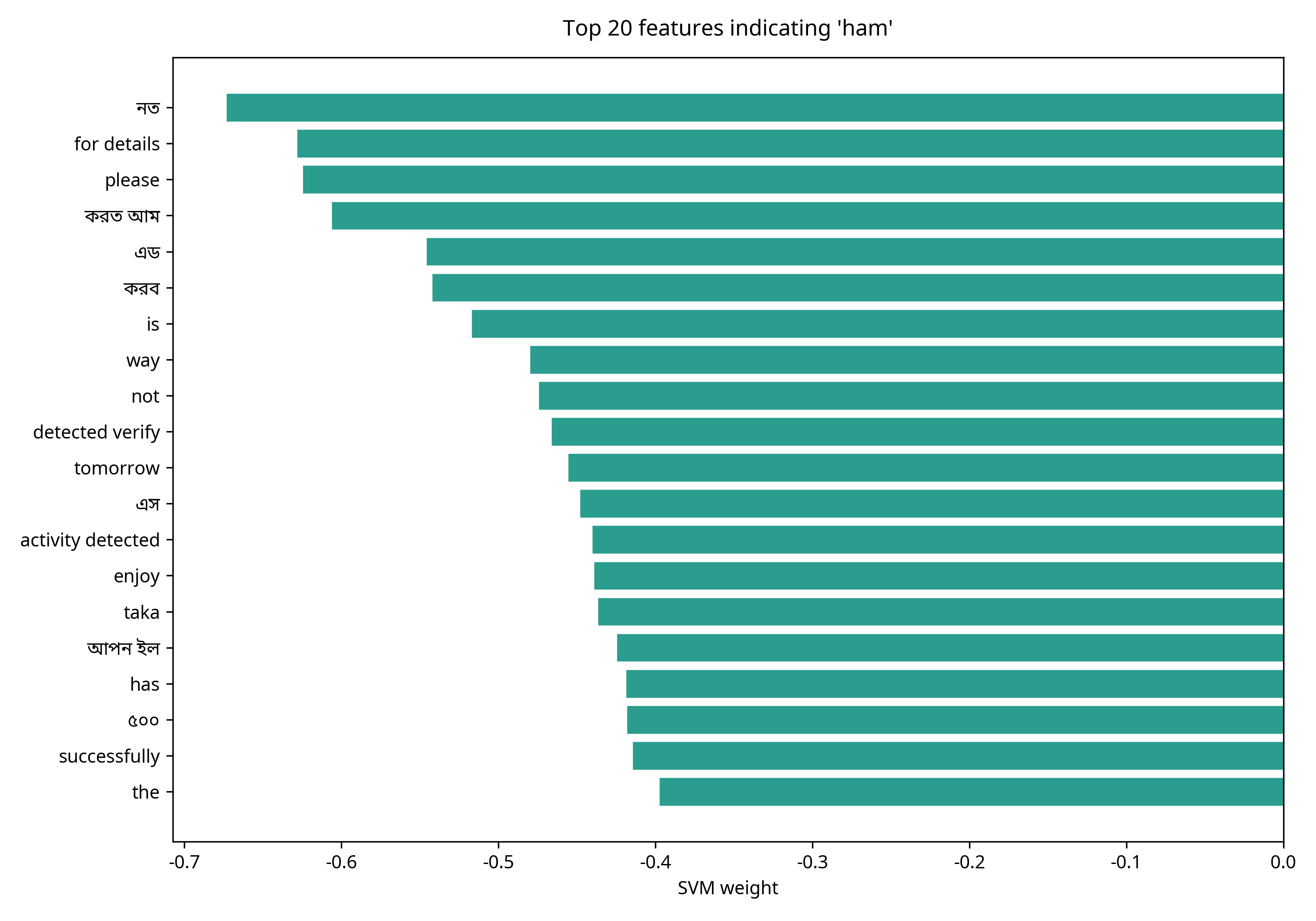}}
\caption{Top 20 TF-IDF features with lowest SVM weights indicating the ham class.}
\label{fig:top_ham}
\end{figure}

Fig.~\ref{fig:top_scam} displays features most indicative of scam messages, including urgency-inducing terms (``now,'' ``urgent,'' ``renew now'') and action-oriented phrases (``claim your,'' ``verify''), along with Bangla promotional terms. Conversely, Fig.~\ref{fig:top_ham} shows ham indicators such as transactional confirmations (``successfully,'' ``activity detected'') and currency references (``taka''), suggesting legitimate messages contain specific transaction details rather than vague promotions.

\subsubsection{Structural Indicators}
\begin{figure}[htbp]
\centerline{\includegraphics[width=\columnwidth]{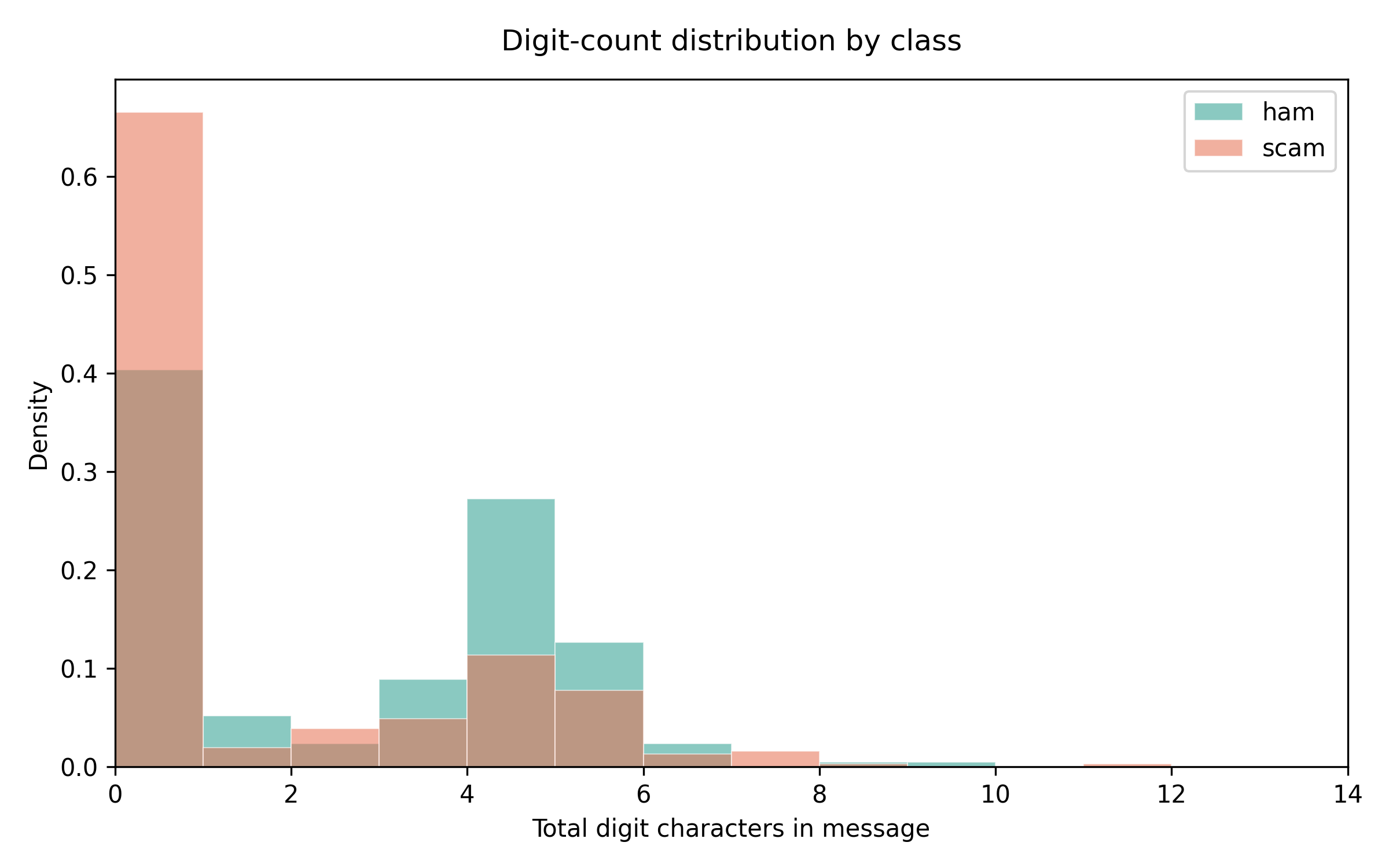}}
\caption{Distribution of digit characters in messages by class.}
\label{fig:digit_count}
\end{figure}

\begin{figure}[htbp]
\centerline{\includegraphics[width=\columnwidth]{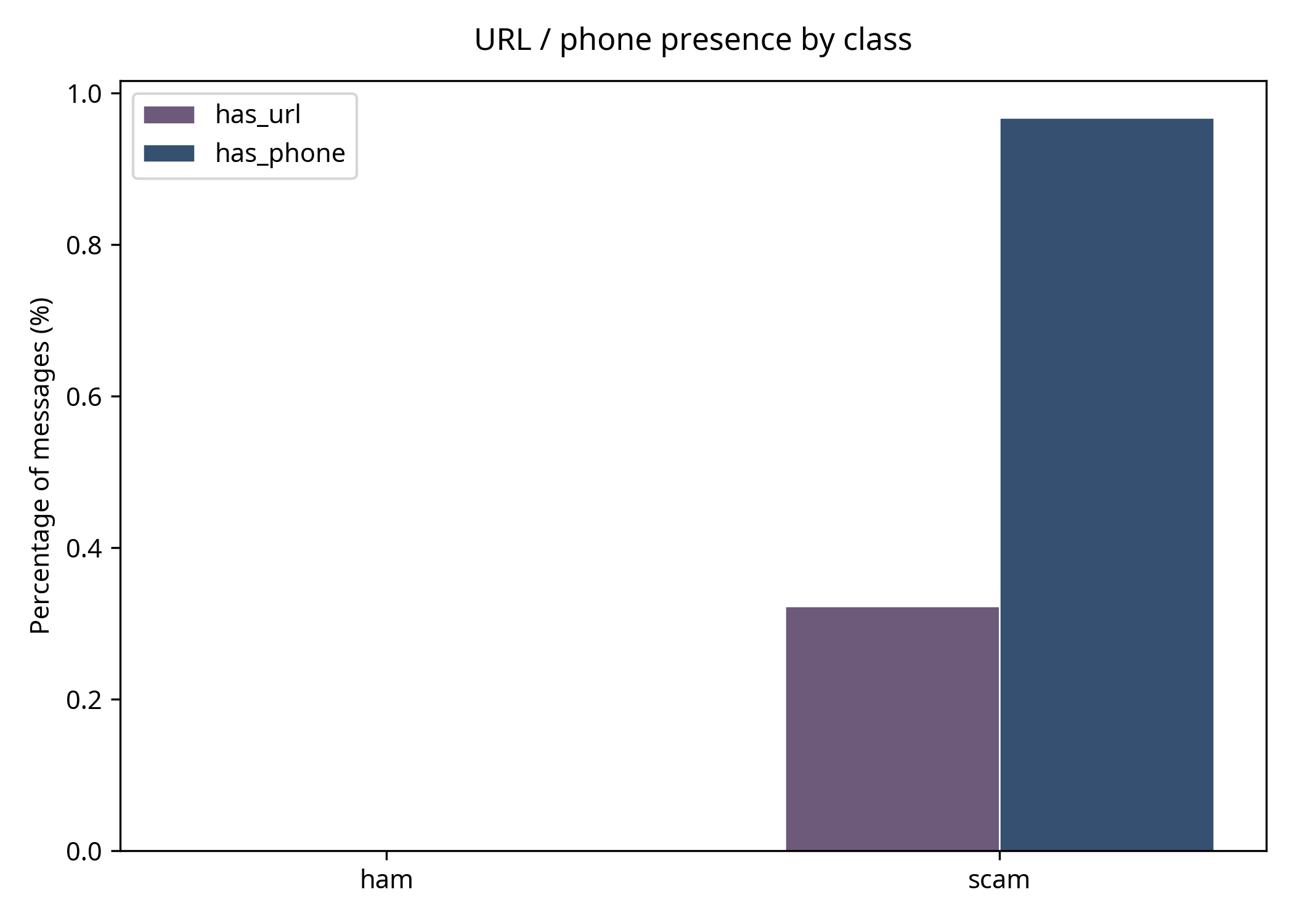}}
\caption{Percentage of messages containing URLs and phone numbers by class.}
\label{fig:url_phone}
\end{figure}

Fig.~\ref{fig:digit_count} shows legitimate messages exhibit a bimodal digit distribution (peaks at zero and four digits), while scam messages show dispersed patterns with frequent numerical elements. Fig.~\ref{fig:url_phone} reveals a striking distinction: legitimate messages contain virtually no URLs or phone numbers, whereas approximately 32\% of scam messages include URLs and 97\% contain phone numbers---strong indicators of fraudulent intent.

These observations demonstrate clear distributional and linguistic differences between classes, motivating the feature representations and classification approaches evaluated subsequently.

% ============================================
% EXPERIMENTAL RESULTS SECTION
% ============================================

\subsection{Experimental Results}

This section presents the experimental evaluation of both classical machine learning models and transformer-based architectures for multilingual financial fraud detection. All models are evaluated using 5-fold stratified cross-validation to ensure robust performance estimation across different data partitions. We report accuracy, macro-averaged F1 score, and Precision-Recall Area Under Curve (PR-AUC) as primary evaluation metrics.

\subsubsection{Overall Performance Comparison}

Table~\ref{tab:overall_results} summarizes the cross-validation performance of all evaluated models. Among classical machine learning approaches, Linear SVM achieves the highest accuracy (91.59\%) and F1 score (91.30\%), followed closely by the Ensemble method (91.21\% accuracy, 90.91\% F1) and Logistic Regression (91.02\% accuracy, 90.73\% F1). The transformer-based model achieves competitive performance with 89.49\% accuracy and 88.88\% F1 score, though with slightly higher variance across folds.

\begin{table}[htbp]
\centering
\caption{Cross-Validation Performance Comparison (Mean $\pm$ Std \%)}
\label{tab:overall_results}
\resizebox{\columnwidth}{!}{%
\begin{tabular}{lccc}
\hline
\textbf{Model} & \textbf{Accuracy (\%)} & \textbf{F1 Macro (\%)} & \textbf{PR-AUC (\%)} \\
\hline
Logistic Regression & 91.02 $\pm$ 2.84 & 90.73 $\pm$ 2.85 & 96.82 $\pm$ 1.52 \\
Linear SVM & \textbf{91.59 $\pm$ 2.19} & \textbf{91.30 $\pm$ 2.18} & 97.17 $\pm$ 1.57 \\
Ensemble & 91.21 $\pm$ 2.49 & 90.91 $\pm$ 2.48 & \textbf{97.19 $\pm$ 1.49} \\
Transformer & 89.49 $\pm$ 2.31 & 88.88 $\pm$ 2.66 & 95.78 $\pm$ 1.09 \\
\hline
\end{tabular}%
}
\end{table}

Notably, all models achieve PR-AUC scores exceeding 95\%, indicating strong ranking performance for distinguishing fraudulent from legitimate messages. The Ensemble model achieves the highest PR-AUC (97.19\%), suggesting robust probability calibration across decision thresholds.

\subsubsection{Confusion Matrix Analysis}

\begin{figure}[htbp]
\centering
\includegraphics[width=\columnwidth]{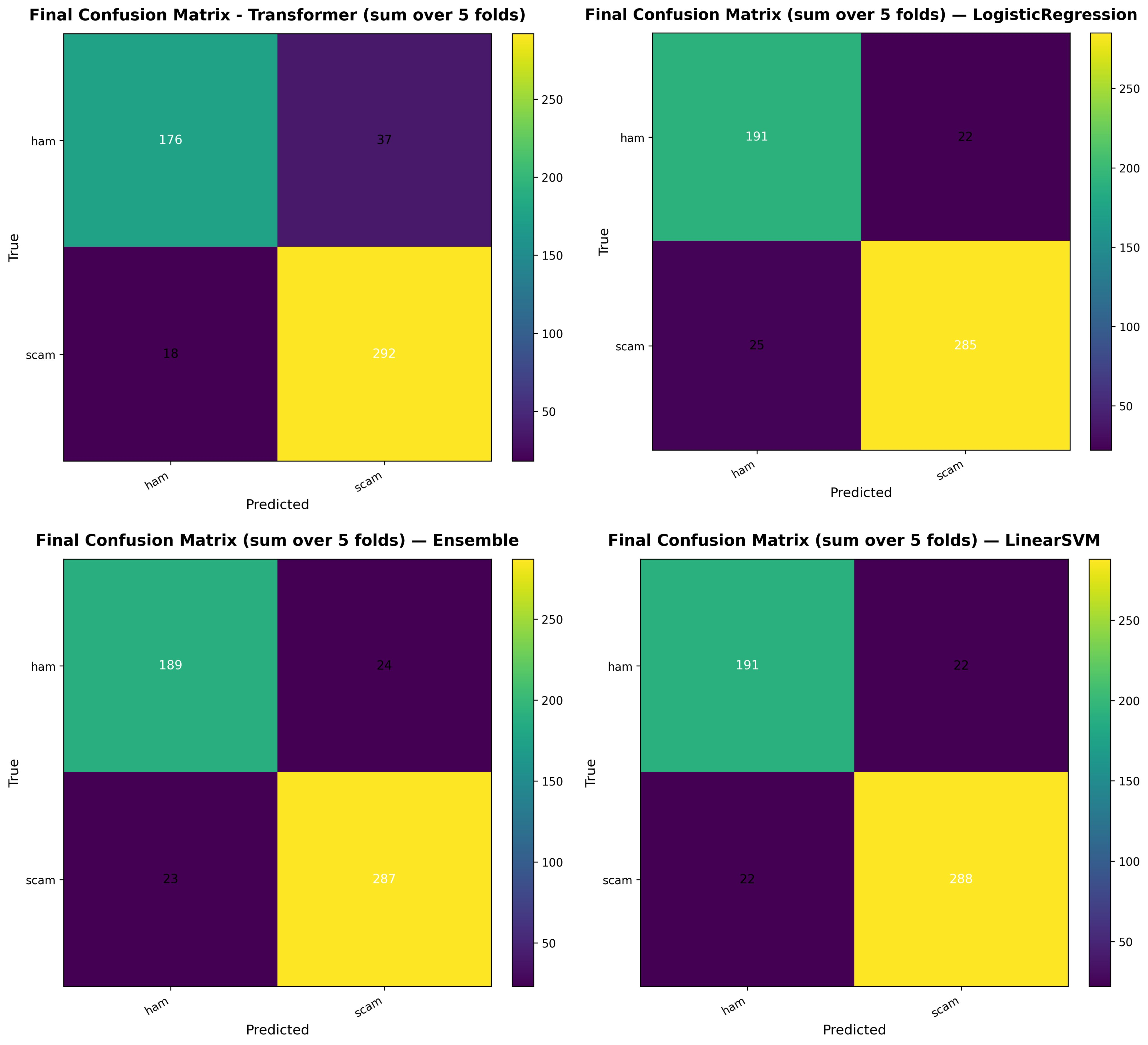}
\caption{Aggregated confusion matrices (sum over 5 folds) for Transformer, Logistic Regression, Ensemble, and Linear SVM models. Rows represent true labels and columns represent predicted labels.}
\label{fig:confusion_matrices}
\end{figure}

Fig.~\ref{fig:confusion_matrices} presents the aggregated confusion matrices (summed over 5 folds) for all evaluated models. The confusion matrix analysis reveals distinct error patterns across different modeling approaches.

Table~\ref{tab:confusion_summary} summarizes the confusion matrix values and derived per-class metrics. Several key observations emerge from this analysis:

\begin{table}[htbp]
\centering
\caption{Confusion Matrix Summary and Per-Class Performance (Aggregated over 5 Folds)}
\label{tab:confusion_summary}
\resizebox{\columnwidth}{!}{%
\begin{tabular}{lcccccc}
\hline
\textbf{Model} & \textbf{TN} & \textbf{FP} & \textbf{FN} & \textbf{TP} & \textbf{Scam F1 (\%)} & \textbf{Ham F1 (\%)} \\
\hline
Transformer & 176 & 37 & 18 & 292 & 91.39 & 86.49 \\
Logistic Regression & 191 & 22 & 25 & 285 & 92.38 & 89.04 \\
Linear SVM & 191 & 22 & 22 & 288 & \textbf{92.90} & \textbf{89.67} \\
Ensemble & 189 & 24 & 23 & 287 & 92.43 & 88.94 \\
\hline
\end{tabular}%
}
\end{table}

\textbf{Scam Detection (Positive Class):} The transformer model achieves the highest recall for detecting fraudulent messages (94.19\%, with only 18 false negatives), indicating strong sensitivity to fraud patterns. However, this comes at the cost of lower precision (88.75\%) due to a higher false positive rate (37 legitimate messages misclassified as scam). In contrast, Linear SVM and Logistic Regression achieve more balanced precision-recall trade-offs for the scam class, with F1 scores of 92.90\% and 92.38\%, respectively.

\textbf{Ham Detection (Negative Class):} Classical machine learning models substantially outperform the transformer in identifying legitimate messages. Linear SVM achieves the highest ham class F1 score (89.67\%), correctly classifying 191 out of 213 legitimate messages. The transformer model shows a notable weakness in ham detection (F1 = 86.49\%), misclassifying 37 legitimate messages as fraudulent---nearly double the false positive rate of Linear SVM (22 false positives).

\textbf{Overall Error Distribution:} The total misclassification count across 523 test samples is lowest for Linear SVM (44 errors), followed by Logistic Regression and Ensemble (47 errors each), and highest for the transformer model (55 errors). This pattern suggests that TF-IDF features combined with linear classifiers provide more discriminative representations for this multilingual fraud detection task compared to the transformer architecture.

\subsubsection{Discussion}

The experimental results reveal several important findings regarding multilingual financial fraud detection:

\textbf{Classical ML Competitiveness:} Despite the recent dominance of transformer architectures in NLP tasks, classical machine learning models with TF-IDF features achieve superior performance on this multilingual fraud detection dataset. Linear SVM achieves the best overall accuracy (91.59\%) and most balanced per-class performance, outperforming the transformer model by approximately 2 percentage points in accuracy.

\textbf{Transformer Trade-offs:} The transformer model exhibits a bias toward predicting the scam class, resulting in high fraud recall but elevated false positive rates. This behavior may be advantageous in high-stakes fraud detection scenarios where missing fraudulent messages carries greater cost than occasional false alarms. However, the higher false positive rate could lead to user frustration if legitimate financial communications are frequently flagged.

\textbf{Stability Considerations:} Classical models demonstrate lower variance across cross-validation folds compared to the transformer, suggesting more stable performance across different data partitions. This stability is particularly valuable for deployment in production systems where consistent performance is required.

\textbf{Multilingual Challenges:} The relatively modest performance gap between models suggests that the multilingual and code-mixed nature of the dataset poses challenges for all approaches. The presence of Bangla text, English text, and code-mixed content may limit the effectiveness of both TF-IDF representations (which treat languages independently) and multilingual transformers (which may have limited exposure to Bangla financial vocabulary during pretraining).

\section{Conclusion}
This study investigated financial fraud detection in a multilingual Bangla–English setting, comparing classical machine learning approaches with transformer-based architectures using 5-fold stratified cross-validation. Classical machine learning models with TF-IDF features consistently outperformed transformer architectures, with Linear SVM achieving the highest accuracy (91.59\%) and F1 score (91.30\%), whereas the transformer model attained 89.49\% accuracy with superior fraud recall (94.19\%) but nearly double the false positive rate. All models achieved PR-AUC scores exceeding 95\%, indicating strong ranking performance. Exploratory analysis revealed distinctive characteristics of fraudulent messages: longer length, urgency-inducing terminology, and the predominant inclusion of URLs (32\%) and phone numbers (97\%)—structural features virtually absent in legitimate messages, validating TF-IDF effectiveness in capturing fraud-indicative patterns. Key challenges identified include linguistic diversity (80\% English, 15–17\% Bangla, and code-mixed content), limited transformer exposure to domain-specific Bangla financial vocabulary, and the need for language-agnostic representations. Future work should explore fine-tuned multilingual models, hybrid approaches combining TF-IDF with transformer embeddings, and larger annotated datasets for low-resource language fraud detection. This study demonstrates that classical approaches remain competitive alternatives to deep learning in resource-constrained multilingual settings.

\bibliographystyle{IEEEtran}
\bibliography{references}

@article{salton1988term,
  title={Term-weighting approaches in automatic text retrieval},
  author={Salton, Gerard and Buckley, Christopher},
  journal={Information processing \& management},
  volume={24},
  number={5},
  pages={513--523},
  year={1988},
  publisher={Elsevier}
}

@inproceedings{joachims1998text,
  title={Text categorization with support vector machines: Learning with many relevant features},
  author={Joachims, Thorsten},
  booktitle={European conference on machine learning},
  pages={137--142},
  year={1998},
  organization={Springer}
}

@article{fan2008liblinear,
  title={LIBLINEAR: A library for large linear classification},
  author={Fan, Rong-En and Chang, Kai-Wei and Hsieh, Cho-Jui and Wang, Xiang-Rui and Lin, Chih-Jen},
  journal={the Journal of machine Learning research},
  volume={9},
  pages={1871--1874},
  year={2008},
  publisher={JMLR. org}
}

@article{vaswani2017attention,
  title={Attention is all you need},
  author={Vaswani, Ashish and Shazeer, Noam and Parmar, Niki and Uszkoreit, Jakob and Jones, Llion and Gomez, Aidan N and Kaiser, {\L}ukasz and Polosukhin, Illia},
  journal={Advances in neural information processing systems},
  volume={30},
  year={2017}
}

@inproceedings{devlin2019bert,
  title={Bert: Pre-training of deep bidirectional transformers for language understanding},
  author={Devlin, Jacob and Chang, Ming-Wei and Lee, Kenton and Toutanova, Kristina},
  booktitle={Proceedings of the 2019 conference of the North American chapter of the association for computational linguistics: human language technologies, volume 1 (long and short papers)},
  pages={4171--4186},
  year={2019}
}

@article{amin2022detecting,
  title={Detecting conspiracy theory against covid-19 vaccines},
  author={Amin, Md Hasibul and Madanu, Harika and Lavu, Sahithi and Mansourifar, Hadi and Alsagheer, Dana and Shi, Weidong},
  journal={arXiv preprint arXiv:2211.13003},
  year={2022}
}

@inproceedings{joshi2020state,
  title={The state and fate of linguistic diversity and inclusion in the NLP world},
  author={Joshi, Pratik and Santy, Sebastin and Budhiraja, Amar and Bali, Kalika and Choudhury, Monojit},
  booktitle={Proceedings of the 58th annual meeting of the association for computational linguistics},
  pages={6282--6293},
  year={2020}
}

@article{hossain2024advancements,
  title={Advancements in natural Language processing: Leveraging transformer models for multilingual text Generation},
  author={Hossain, Mohammad Zobair and Goyal, Sachin},
  journal={Pacific Journal of Advanced Engineering Innovations},
  volume={1},
  number={1},
  pages={4--12},
  year={2024}
}

@article{alam2021review,
  title={A review of bangla natural language processing tasks and the utility of transformer models},
  author={Alam, Firoj and Hasan, Arid and Alam, Tanvirul and Khan, Akib and Tajrin, Janntatul and Khan, Naira and Chowdhury, Shammur Absar},
  journal={arXiv preprint arXiv:2107.03844},
  year={2021}
}

@article{rahman2019benchmark,
  title={A Benchmark Dataset for Bangla Text Classification},
  author={Rahman, Md. Mizanur and others},
  journal={IEEE Access},
  volume={7},
  pages={52585--52599},
  year={2019}
}

@article{ngai2011application,
  title={The application of data mining techniques in financial fraud detection: A classification framework and an academic review of literature},
  author={Ngai, Eric WT and Hu, Yong and Wong, Yiu Hing and Chen, Yijun and Sun, Xin},
  journal={Decision support systems},
  volume={50},
  number={3},
  pages={559--569},
  year={2011},
  publisher={Elsevier}
}

@article{bahnsen2015cost,
  title={Cost-sensitive decision trees for fraud detection},
  author={Bahnsen, AC and Aouada, D and Ottersten, B},
  journal={Expert Systems with Applications},
  volume={42},
  number={22},
  pages={1--13},
  year={2015}
}

@inproceedings{artetxe2020cross,
  title={On the cross-lingual transferability of monolingual representations},
  author={Artetxe, Mikel and Ruder, Sebastian and Yogatama, Dani},
  booktitle={Proceedings of the 58th annual meeting of the association for computational linguistics},
  pages={4623--4637},
  year={2020}
}

@article{dal2015adversarial,
  title={Calibrating Probability with Undersampling for Unbalanced Classification},
  author={Dal Pozzolo, Andrea and Caelen, Olivier and Le Borgne, Yann-A{\"e}l and Waterschoot, Serge and Bontempi, Gianluca},
  journal={IEEE Symposium Series on Computational Intelligence},
  pages={159--166},
  year={2015}
}

@article{carcillo2018scarff,
  title={Scarff: a scalable framework for streaming credit card fraud detection with spark},
  author={Carcillo, Fabrizio and Dal Pozzolo, Andrea and Le Borgne, Yann-A{\"e}l and Caelen, Olivier and Mazzer, Yannis and Bontempi, Gianluca},
  journal={Information fusion},
  volume={41},
  pages={182--194},
  year={2018},
  publisher={Elsevier}
}

@misc{ahmed2025financial,
  author       = {Ahmed, Al Rafi and Islam, Gazi Faizul},
  title        = {Financial Scams Detection Dataset},
  year         = {2025},
  publisher    = {Mendeley Data},
  version      = {V2},
  doi          = {10.17632/znsk27yk3h.2}
}

@inproceedings{aguilar2020lince,
  title={LinCE: A centralized benchmark for linguistic code-switching evaluation},
  author={Aguilar, Gustavo and Kar, Sudipta and Solorio, Thamar},
  booktitle={Proceedings of the Twelfth Language Resources and Evaluation Conference},
  pages={1803--1813},
  year={2020}
}

@article{phua2010comprehensive,
  title={A comprehensive survey of data mining-based fraud detection research},
  author={Phua, Clifton and Lee, Vincent and Smith, Kate and Gayler, Ross},
  journal={arXiv preprint arXiv:1009.6119},
  year={2010}
}

@article{dal2017credit,
  title={Credit card fraud detection: a realistic modeling and a novel learning strategy},
  author={Dal Pozzolo, Andrea and Boracchi, Giacomo and Caelen, Olivier and Alippi, Cesare and Bontempi, Gianluca},
  journal={IEEE transactions on neural networks and learning systems},
  volume={29},
  number={8},
  pages={3784--3797},
  year={2017},
  publisher={IEEE}
}

@article{khanuja2021muril,
  title={Muril: Multilingual representations for indian languages},
  author={Khanuja, Simran and Bansal, Diksha and Mehtani, Sarvesh and Khosla, Savya and Dey, Atreyee and Gopalan, Balaji and Margam, Dilip Kumar and Aggarwal, Pooja and Nagipogu, Rajiv Teja and Dave, Shachi and others},
  journal={arXiv preprint arXiv:2103.10730},
  year={2021}
}

@inproceedings{kakwani2020indicnlpsuite,
  title={IndicNLPSuite: Monolingual corpora, evaluation benchmarks and pre-trained multilingual language models for Indian languages},
  author={Kakwani, Divyanshu and Kunchukuttan, Anoop and Golla, Satish and NC, Gokul and Bhattacharyya, Avik and Khapra, Mitesh M and Kumar, Pratyush},
  booktitle={Findings of the association for computational linguistics: EMNLP 2020},
  pages={4948--4961},
  year={2020}
}

@inproceedings{pires2019multilingual,
  title={How multilingual is multilingual BERT?},
  author={Pires, Telmo and Schlinger, Eva and Garrette, Dan},
  booktitle={Proceedings of the 57th annual meeting of the association for computational linguistics},
  pages={4996--5001},
  year={2019}
}

@article{artetxe2019massively,
  title={Massively multilingual sentence embeddings for zero-shot cross-lingual transfer and beyond},
  author={Artetxe, Mikel and Schwenk, Holger},
  journal={Transactions of the association for computational linguistics},
  volume={7},
  pages={597--610},
  year={2019},
  publisher={MIT Press One Rogers Street, Cambridge, MA 02142-1209, USA journals-info~…}
}

@inproceedings{conneau2020unsupervised,
  title={Unsupervised cross-lingual representation learning at scale},
  author={Conneau, Alexis and Khandelwal, Kartikay and Goyal, Naman and Chaudhary, Vishrav and Wenzek, Guillaume and Guzm{\'a}n, Francisco and Grave, Edouard and Ott, Myle and Zettlemoyer, Luke and Stoyanov, Veselin},
  booktitle={Proceedings of the 58th annual meeting of the association for computational linguistics},
  pages={8440--8451},
  year={2020}
}

@article{olowookere2020framework,
  title={A framework for detecting credit card fraud with cost-sensitive meta-learning ensemble approach},
  author={Olowookere, Toluwase Ayobami and Adewale, Olumide Sunday},
  journal={Scientific African},
  volume={8},
  pages={e00464},
  year={2020},
  publisher={Elsevier}
}

@article{boulieris2024fraud,
  title={Fraud detection with natural language processing},
  author={Boulieris, Petros and Pavlopoulos, John and Xenos, Alexandros and Vassalos, Vasilis},
  journal={Machine Learning},
  volume={113},
  number={8},
  pages={5087--5108},
  year={2024},
  publisher={Springer}
}

@inproceedings{rodriguez2022natural,
  title={A natural language processing approach for financial fraud detection},
  author={Rodr{\'\i}guez, Javier Fern{\'a}ndez and Papale, Michele and Carminati, Michele and Zanero, Stefano and others},
  booktitle={CEUR workshop proceedings},
  volume={3260},
  pages={135--149},
  year={2022},
  organization={CEUR-WS. org}
}

@article{craja2020deep,
  title={Deep learning for detecting financial statement fraud},
  author={Craja, Patricia and Kim, Alisa and Lessmann, Stefan},
  journal={Decision Support Systems},
  volume={139},
  pages={113421},
  year={2020},
  publisher={Elsevier}
}

@article{jain2024knowledge,
  title={Knowledge-based data processing for multilingual natural language analysis},
  author={Jain, Deepak Kumar and Eyre, Yamila Garc{\'\i}a-Mart{\'\i}nez and Kumar, Akshi and Gupta, Brij B and Kotecha, Ketan},
  journal={ACM Transactions on Asian and Low-Resource Language Information Processing},
  volume={23},
  number={5},
  pages={1--16},
  year={2024},
  publisher={ACM New York, NY}
}

@article{paper1,
  doi = {10.13140/RG.2.2.32561.04960/2},
  url = {https://www.researchgate.net/doi/10.13140/RG.2.2.32561.04960/2},
  author = {{Hasibul Amin} and Madanu, Harika and {Sahithi Lavu} and Mansourifar, Hadi and Alsagheer, Dana and {Weidong Shi}
  },
  language = {en},
  title = {Detecting Conspiracy Theory Against COVID-19 Vaccines},
  publisher = {Unpublished},
  year = {2022},
  Journal= {arXiv preprint arXiv:2211.13003}
}

@INPROCEEDINGS{11103694,
  author={Ahmed, Tanvir and Ferdus, Mst Jannatul and Parvez, Rezwanul and Lachchu, Md Amzad Hossain and Arefin, Sydul and Ahsan, Mostofa},
  booktitle={2025 IEEE International Conference on Electro Information Technology (eIT)}, 
  title={Financial Fraud Detection using Synthetic Dataset: An Analysis of Machine Learning Algorithms}, 
  year={2025},
  volume={},
  number={},
  pages={406-411},
  doi={10.1109/eIT64391.2025.11103694}}

@phdthesis{ahmed2022sentiment,
  author   = {Ahmed, Tanvir},
  title    = {Sentiment Analysis of {COVID-19} Vaccination Impact on Twitter Tweets Using {NLP} Supervised Learning and {RNN} Classification Comparison},
  year     = {2022},
  school   = {ProQuest Dissertations and Theses},
  pages    = {51},
  url      = {https://www.proquest.com/dissertations-theses/sentiment-analysis-covid-19-vaccination-impact-on/docview/2777441422/se-2},
  isbn     = {9798371995292}
}

@INPROCEEDINGS{10459734,
  author={Hasan, Munjur and Rahman, Md Saifur and Islam, Sabrina and Ahmed, Tanvir and Rifat, Nafiz and Ahsan, Mostofa and Gomes, Rahul and Chowdhury, Md.},
  booktitle={2023 International Conference on Machine Learning and Applications (ICMLA)}, 
  title={Vision Transformer-based Classification for Lung and Colon Cancer using Histopathology Images}, 
  year={2023},
  volume={},
  number={},
  pages={1300-1304},
  doi={10.1109/ICMLA58977.2023.00196}}

@Article{electronics14030554,
AUTHOR = {Zidan, Arif Hassan and Jahin, Afrar and Bao, Yu and Zhang, Wei},
TITLE = {Semi-Nonlinear Deep Efficient Reconstruction for Unveiling Linear and Nonlinear Spatial Features of the Human Brain},
JOURNAL = {Electronics},
VOLUME = {14},
YEAR = {2025},
NUMBER = {3},
ARTICLE-NUMBER = {554},
URL = {https://www.mdpi.com/2079-9292/14/3/554},
ISSN = {2079-9292},
DOI = {10.3390/electronics14030554}
}

@article{ZIDAN20251128,
title = {Mapping longitudinally consistent intrinsic connectivity networks in macaque brain via longitudinal sparse dictionary learning},
journal = {IBRO Neuroscience Reports},
volume = {19},
pages = {1128-1140},
year = {2025},
issn = {2667-2421},
doi = {https://doi.org/10.1016/j.ibneur.2024.11.014},
url = {https://www.sciencedirect.com/science/article/pii/S2667242124001076},
author = {Arif Hassan Zidan and Afrar Jahin and Yu Bao and Wei Zhang},
}

@inproceedings{sabuj2024recondite,
  title={Recondite thyroid pathology prediction: Hermeneutic integration of neural and machine learning architectures},
  author={Sabuj, Md Sanowar Hossain and Imam, Touhid and Islam, Jahirul and Sultana, Sharmin and Uddin, Mohammad Shihab and Uddin, Bushra},
  booktitle={2024 IEEE International Conference on Computing (ICOCO)},
  pages={267--272},
  year={2024},
  organization={IEEE}
}

@inproceedings{uddin2024epistemological,
  title={Epistemological Advancements in Cardiological Forecasting: Machine Learning as a Paradigm for Prognostic Precision},
  author={Uddin, Bushra and Uddin, Mohammad Shihab and Sultana, Sharmin and Sabuj, MD Sanowar Hossain and Neha, Fariha Ferdous and Uddin, MD Salah},
  booktitle={2024 International Conference on Computer and Applications (ICCA)},
  pages={01--06},
  year={2024},
  organization={IEEE}
}

@inproceedings{islam2024hierarchical,
  title={Hierarchical superiority of machine learning ensembles in diabetes risk apprehension: A granular analytical perspective},
  author={Islam, Jahirul and Imam, Touhid and Sultana, Sharmin and Sabuj, Md Sanowar Hossain and Uddin, Mohammad Shihab and Uddin, Bushra},
  booktitle={2024 IEEE International Conference on Computing (ICOCO)},
  pages={255--260},
  year={2024},
  organization={IEEE}
}

@inproceedings{sultana2024deciphering,
  title={Deciphering Oncological Recurrence through Hybridized Machine Learning Paradigms in Thyroid Disorders},
  author={Sultana, Sharmin and Uddin, Salah and Uddin, Bushra and Uddin, Mohammad Shihab and Neha, Fariha Ferdous},
  booktitle={2024 International Conference on Computer and Applications (ICCA)},
  pages={1--6},
  year={2024},
  organization={IEEE}
}

@inproceedings{sultana2024analyzing,
  title={Analyzing Neuroimaging Epiphenomena: Machine Learning Approaches in Alzheimer's Prognostication},
  author={Sultana, Sharmin and Uddin, Bushra and Uddin, Mohammad Shihab and Uddin, MD Salah and Karim, Fazle and Mehedi, Mohiuddin},
  booktitle={2024 International Conference on Computer and Applications (ICCA)},
  pages={1--6},
  year={2024},
  organization={IEEE}
}

@inproceedings{imam2024ml,
  title={ML-Driven Solutions for Chronic Liver Disease: Predictive Models and Prevention Strategies},
  author={Imam, Touhid and Islam, Jahirul and Sultana, Sharmin and Uddin, Md Salah and Uddin, Mohammad Shihab and Uddin, Bushra},
  booktitle={2024 IEEE International Conference on Computing (ICOCO)},
  pages={261--266},
  year={2024},
  organization={IEEE}
}

@inproceedings{raju2024ontological,
  title={An ontological framework for lung carcinoma prognostication via sophisticated stacking and synthetic minority oversampling techniques},
  author={Raju, Md Azad Hossain and Imam, Touhid and Islam, Jahirul and Al Rakin, Abdullah and Nayyem, Mohammad Navid and Uddin, Mohammad Shihab},
  booktitle={2024 IEEE Asia Pacific conference on wireless and mobile (APWiMob)},
  pages={125--130},
  year={2024},
  organization={IEEE}
}

@inproceedings{wang2025explainability,
  title={Explainability-driven defense: Grad-cam-guided model refinement against adversarial threats},
  author={Wang, Longwei and Uddin, Ifrat Ikhtear and Qin, Xiao and Zhou, Yang and Santosh, KC},
  booktitle={Proceedings of the AAAI Symposium Series},
  volume={6},
  number={1},
  pages={49--57},
  year={2025}

}

@inproceedings{arefin2024retail,
  title={Retail industry analytics: Unraveling consumer behavior through rfm segmentation and machine learning},
  author={Arefin, Sydul and Parvez, Rezwanul and Ahmed, Tanvir and Ahsan, Mostofa and Sumaiya, Fnu and Jahin, Fariha and Hasan, Munjur},
  booktitle={2024 IEEE international conference on electro information technology (eIT)},
  pages={545--551},
  year={2024},
  organization={IEEE}
}

@inproceedings{arefin2024enhancing,
  title={Enhancing bank consumer retention: a comprehensive analysis of churn prediction using machine-learning and deep learning techniques},
  author={Arefin, Sydul and Parvez, Rezwanul and Ahmed, Tanvir and Jahin, Fariha and Sumaiya, Fnu and Ahsan, Mostofa},
  booktitle={International Conference on Advances in Information Communication Technology \& Computing},
  pages={265--281},
  year={2024},
  organization={Springer}
}

@inproceedings{parvez2025can,
  title={How can Big Data Enhance Machine Learning Model Performance in Predicting Future Prices? The Case of the US Consumer Price Index},
  author={Parvez, Rezwanul and Ferdus, Mst Jannatul and Lachchu, Md Amzad Hossain and Ahmed, Tanvir and Arefin, Sydul and Ahsan, Mostofa},
  booktitle={2025 IEEE International Conference on Electro Information Technology (eIT)},
  pages={412--418},
  year={2025},
  organization={IEEE}
}

@inproceedings{ferdus2025does,
  title={Does Tweets Play a Role in Shaping Human Sentiment: A Stock Market Analysis Using Machine Learning, Deep Learning, and Pre-Trained Methods},
  author={Ferdus, Mst Jannatul and Parvez, Rezwanul and Lachchu, Md Amzad Hossain and Ahmed, Tanvir and Arefin, Sydul and Ahsan, Mostofa},
  booktitle={2025 IEEE International Conference on Electro Information Technology (eIT)},
  pages={431--436},
  year={2025},
  organization={IEEE}
}

@inproceedings{ferdus2025stock,
  title={Stock Investor's Sentiment Analysis Using Advanced Machine Learning Algorithms},
  author={Ferdus, Mst Jannatul and Parvez, Rezwanul and Lachchu, Md Amzad Hossain and Ahmed, Tanvir and Arefin, Sydul and Ahsan, Mostofa},
  booktitle={2025 IEEE International Conference on Electro Information Technology (eIT)},
  pages={425--430},
  year={2025},
  organization={IEEE}
}

\end{document}